\documentclass[runningheads]{llncs}
\usepackage{graphicx}

\usepackage{amsmath,amsfonts,bm}

\def\Figref#1{Figure~\ref{#1}}

\def\Secref#1{Section~\ref{#1}}

\def\eqref#1{equation~\ref{#1}}
\def\Eqref#1{Equation~\ref{#1}}

\def\1{\bm{1}}

\def\vg{{\bm{g}}}
\def\vh{{\bm{h}}}

\def\vm{{\bm{m}}}

\def\vq{{\bm{q}}}

\DeclareMathAlphabet{\mathsfit}{\encodingdefault}{\sfdefault}{m}{sl}
\SetMathAlphabet{\mathsfit}{bold}{\encodingdefault}{\sfdefault}{bx}{n}

\def\gA{{\mathcal{A}}}

\def\gE{{\mathcal{E}}}

\def\gG{{\mathcal{G}}}

\def\gN{{\mathcal{N}}}
\def\gO{{\mathcal{O}}}

\def\gR{{\mathcal{R}}}

\def\gV{{\mathcal{V}}}

\def\sR{{\mathbb{R}}}

\newcommand{\softmax}{\mathrm{softmax}}

\usepackage{url}
\usepackage{hyperref}
\usepackage{amsmath}
\usepackage{booktabs}
\usepackage{xspace}
\usepackage{pifont}
\usepackage{threeparttable}
\usepackage{tikz}
\usepackage{graphicx} 
\usepackage{pgfplots}
\usepackage{multirow}
\RequirePackage{algorithm}
\RequirePackage{algorithmic}
\usepackage{enumitem}
\usepackage{wrapfig}
\usepackage{subcaption}
\usepackage[normalem]{ulem}
\usepackage{enumitem}

\begin{document}
\title{SemPool: Simple, robust, and interpretable KG pooling for enhancing language models}
\titlerunning{SemPool: KG pooling for enhancing language models}
\author{Costas Mavromatis\inst{1} \and
Petros Karypis \inst{2} \and
George Karypis \inst{1}}
\institute{University of Minnesota \\
\email{\{mavro016, karypis\}@umn.edu} 
\and
University of California San Diego\\
\email{pkarypis@ucsd.edu}
}

\maketitle              %
\begin{abstract}
Knowledge Graph (KG) powered question answering (QA) performs complex reasoning over language semantics as well as knowledge facts. Graph Neural Networks (GNNs) learn to aggregate information from the underlying KG, which is combined with Language Models (LMs) for effective reasoning with the given question. However, GNN-based methods for QA rely on the graph information of the candidate answer nodes, which  limits their effectiveness in more challenging settings where critical answer information is not included in the KG. We propose a simple graph pooling approach that learns useful semantics of the KG that can  aid the LM's reasoning and that its effectiveness is robust under graph perturbations. Our method, termed SemPool, represents KG facts with pre-trained LMs, learns to aggregate their semantic information, and fuses it at different layers of the LM. 
Our experimental results show that SemPool outperforms state-of-the-art GNN-based methods by 2.27\% accuracy points on average when answer information is missing from the KG. In addition, SemPool offers interpretability on what type of graph information is fused at different LM layers.

\end{abstract}

\section{Introduction}
\vspace{-0.1in}
Question answering (QA) is a  complex reasoning task that requires understanding of a given natural language query, as well as domain-specific knowledge. For instance, answering biomedical questions requires understanding of biomedical terms as well as knowledge about biomedicine. Language models (LMs)~\cite{devlin2018bert,raffel2020t5} are pre-trained on large corpora to understand their underlying semantics. Thus, fine-tuning LMs for the given reasoning tasks~\cite{liu2019roberta,khashabi2020unifiedqa} is the dominant paradigm in NLP for QA. 

Despite their success, LMs struggle on intensive reasoning tasks that require good in-domain knowledge~\cite{mallen2022nottrust}.
As a result, recent methods incorporate Knowledge Graphs (KGs) during the QA task~\cite{mihaylov2018knowledgeable,feng2020scalable}, which are graphs that capture factual knowledge explicitly as triplets. Each triplet consists of two entities and their corresponding relation. 
Most successful KG-based methods~\cite{yasunaga2021qagnn} leverage Graph Neural Networks (GNNs)~\cite{velivckovic2017gat}, which have shown remarkable performance at reasoning tasks with graph information~\cite{mavromatis2022rearev,zhu2022neural}. 

Nevertheless, GNNs operate on graph data while LMs use natural language sequences, which makes information exchange between the two modalities challenging. In fact, our empirical findings (\Secref{sec:findings}) suggest that GNNs mainly provide graph statistical information for the QA task~\cite{wang2021gsc} rather than information that grounds the LM's reasoning and is robust under graph perturbations. In addition, the representation space mismatch between graph (KGs are usually represented with external node embeddings) and language (represented with pre-trained LMs) does not aid the information exchange between the two modalities. 

\begin{figure*}[t]
    \centering
    \includegraphics[width=0.8\linewidth]{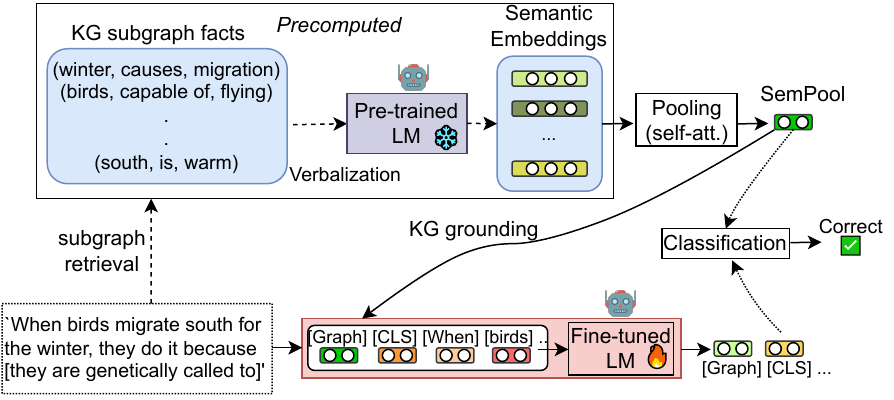}
    \caption{Our SemPool method performs simple graph pooling to enhance the LM's reasoning. Facts of the KG are represented by their semantic information with pre-trained LMs. SemPool aggregates the graph's semantic information into a single representation that is fed into the LM for QA. }
    \vspace{-0.2in}
    \label{fig:method}
\end{figure*}

In this work, we present SemPool, a simple graph pooling method that enhances the LM's reasoning with KG textual information. As illustrated in \Figref{fig:method}, SemPool represents each fact in the KG with the pre-trained LM, aiming at semantic alignment between graph and language.  SemPool then performs a global graph pooling operation in order to aggregate semantic information from the whole graph into a single representation. The aggregated representation is fused as input into the fine-tuned LM for QA, which grounds the LM's reasoning to the information provided. Moreover, we extend SemPool to fuse different type of semantic information into different LM's layers (\Secref{sec:fusion}), providing more flexibility during learning. 

SemPool demonstrates robust performance under different settings. We experiment with standard QA benchmarks (OpenbookQA, RiddleSense, MedQA-USMLE), (i) when complemented by complete in-domain KGs, and (ii) when complemented by in-domain KGs where critical information about the candidate answers is missing. SemPool outperforms the \emph{best} performing GNN-based approach by 2.27\% accuracy points in the challenging case, while it is competitive (second-best) in the easier case. In addition, our experiments show that SemPool is effective under different LMs (\Secref{sec:main_results}), highlight the importance of semantic alignment between language and graph, and illustrate SemPool's interpretability (\Secref{sec:res-abla}). 

\section{Related Work}
\vspace{-0.1in}
\textbf{Question Answering with KGs}. Many KGs have been employed to improve QA for different domains, such as ConceptNet~\cite{speer2017conceptnet} for commonsense QA.
Graph neural networks~\cite{velivckovic2017gat,schlichtkrull2018rgcn} have been widely used to combine KG information with language models~\cite{lin2019kagnet,feng2020scalable,yasunaga2021qagnn,zhang2022greaselm,sun2021jointlk,tian2023graph} leading to SOTA QA systems. In this work, we give new insights on the sensitivity of GNN-based methods with respect to the provided graph and propose a simple approach that improves robustness for QA. Other methods have explored to provide the verbalization of the retrieved KG facts~\cite{agarwal2020tekgen,xie2022unifiedskg} or their embeddings~\cite{park2023qat} as input sequences to the LM, which, however, considerably increases the inference cost due to the extended context. SemPool integrates graph information as a special input token to the LM, offering low computational cost.

\noindent
\textbf{Graph-augmented LMs}. Combining LMs with graphs that include textual features is an emerging research area~\cite{pan2023unifying,liu2023graph-foundation}. Recent methods have explored fine-tuning LMs on graph data~\cite{mavromatis2023grad,zhao2022glem} as well as aiding the pre-training of LMs with graph information~\cite{yasunaga2022linkbert,ioannidis2022efficient,yasunaga2022dragon,xie2023graph}. SemPool provides a new simple approach to fuse graph information into the LM, easily integrated to existing approaches.

\section{Problem Statement \& Preliminaries}
\vspace{-0.1in}

\textbf{Multi-choice QA}. We study the problem of multiple-choice question answering (QA), where given an optional context $c$, a question $q$, and a set of candidate answers $\gA$, the goal is to select the correct answer $a^\ast \in \gA$.
Multiple-choice QA is transformed into a classification problem by  (i) concatenating the question's context with each of the candidate answer $a \in \gA$ into a statement $q_a$, e.g.,  $q_a = [c, q, a] =$ ``\textit{When birds migrate south for the winter, they do it because (A) they are genetically called to}'', and (ii) selecting the most probable statement from $\{[c, q, a]: a \in \gA \}$. Given each input $[c, q, a]$, a fine-tuned LM is used to determine whether the textual input is plausible. The output token representations $\vq^{(L)} = \text{LM}([c, q, a])$ are used for classifying the statement as the correct one, usually followed by a pooling operation. We point the readers for more details to the corresponding papers~\cite{devlin2018bert,liu2019roberta}.

\noindent
\textbf{Knowledge Graphs (KGs)}.
Knowledge Graph (KG) powered QA aims at leveraging external factual information from a KG in order to improve the LM's reasoning ability. For instance, the KG might contain the factual information (\textit{winter, causes, bird migration}), which is relevant for the question $q=$ ``\textit{When birds migrate south for the winter, they do it because?}''.  
Formally, KG is a multi-relational graph $\gG := (\gV, \gE)$ that contains a set nodes (entities) $\gV$ and a set of edges (facts) $\gE$. Set $\gE \subseteq \gV \times \gR \times \gV$ contains facts in the tuple form $(h, r, t)$, where $h,t \in \gV$ and $r \in \gR$ is the relation between nodes $h$ and $t$ ($\gR$ denotes the relation set).

\noindent
\textbf{Subgraph Retrieval}. \label{sec:subgraph}
For each statement $[c, q, a]$, a subgraph $\gG_{q_a} \subseteq \gG$ is retrieved based on the input's context, which may include nodes that correspond to question entities or answer entities. For example,~\cite{yasunaga2021qagnn} performs entity linking between the question's and the KG's entities, and extracts the two-hop nodes between the linked entities, filtering out question-irrelevant edges.  The context-specific subgraph $\gG_{q_a} = (\gV_{q_a}, \gE_{q_a})$ contains a set of nodes $\gV_{q_a}$, the set of relations $\gR$ and a set of facts $\gE_{q_a} \subseteq \gV_{q_a} \times \gR \times \gV_{q_a}$. Note that different candidate answers $a \in \gA$ for question $q$  lead to different subgraphs $\gG_{q_a}$ as the context changes. 
Similar to previous works~\cite{zhang2022greaselm}, a virtual question node is added to each subgraph and is linked to question and answer entities, by adding \textit{(question, entity, birds)} and \textit{(question, a\_entity, children)} to the edge set $\gE_{q_a}$, for example.

\noindent
\textbf{Graph Neural Networks (GNNs)}. \label{sec:gnn}
GNNs learn to update the representation of a node $v$ by aggregating representations of its neighbors, set $\gN(v)$, in a recursive manner. Following the message passing strategy~\cite{gilmer2017neural}, GNNs update the representation $\vh_v^{(l)}$ of node $v$ at layer $l$ as
\begin{equation}
    \vh_v^{(l)} = \psi \Big( \vh_v^{(l-1)}, \phi \big( \{ \vm^{(l)}_{(v,v')}: v' \in \gN(v) \} \big) \Big),
\end{equation}
where $\vm_{(v,v')}$ is the message between two entities $v$ and $v'$, linked with a relation $r$, and depends on their corresponding representations. Function $\phi(\cdot)$ is an aggregation, e.g., sum, of all neighboring messages and function $\psi(\cdot)$ is a neural network. In order to enable language to graph information fusion, many QA GNN-based approaches~\cite{yasunaga2021qagnn,zhang2022greaselm} set the question node's embedding to the question representation obtained by the LM.

\begin{figure*}[t]
    \centering
    \includegraphics[width=\linewidth]{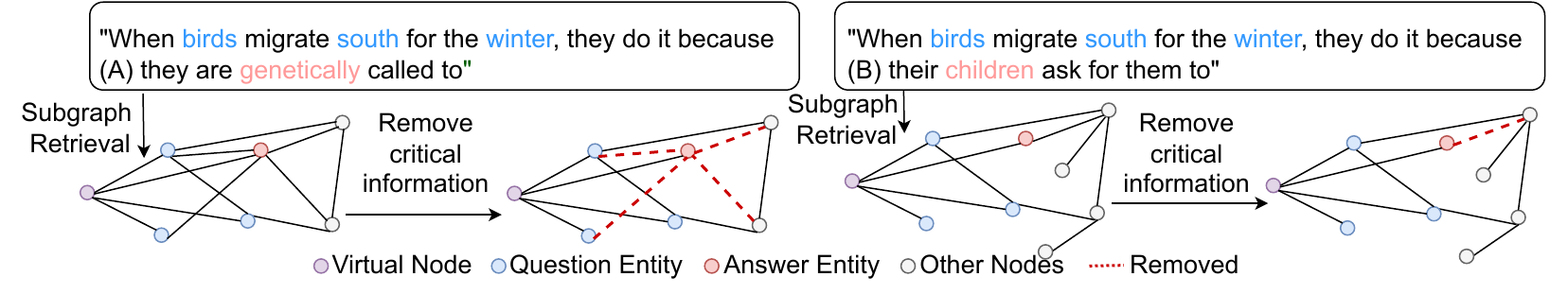}
    \vspace{-0.2in}
    \caption{Setting when critical answer information is removed from the KG. Originally, GNNs propagate information from the answer nodes (\textcolor{pink}{light-red color}) to other nodes of the graph, and the candidate answer with more links is more likely to be the correct answer~\cite{wang2021gsc}. If we remove the answer node's edges, information propagation becomes challenging and GNNs struggle to discriminate between correct and incorrect answers (\Secref{sec:findings}). }
    \vspace{-0.2in}
    \label{fig:graph_adv}
\end{figure*}

\section{Empirical Findings on Robustness} \label{sec:findings}

\vspace{-0.1in}

As discussed in \Secref{sec:gnn}, GNNs leverage the graph information of the retrieved KG to update the node embeddings. However, current QA GNN-based approaches use external node embeddings to represent the nodes' information. The representation space of these embeddings is not necessarily aligned with the representation space of the LM, which limits the effectiveness of fusing semantic information between natural language and graph. For example, \cite{wang2021gsc} and \cite{park2023qat} show that replacing the node embeddings with simple node features, such as node types (node coloring in \Figref{fig:graph_adv}), leads to better QA performance. These findings indicate that  GNNs rely on the underlying graph statistics, e.g., the number of connections between answer nodes and other graph nodes~\cite{wang2021gsc}, to discriminate between correct and incorrect answers.

To test our hypothesis, we experiment with a  setting for multiple-choice QA, where critical information is missing from the KG. For each retrieved KG subgraph, we remove the facts (edges) that include the  candidate answer $a$, similar to the case where answer entities are not linked in the graph. The setting becomes challenging as GNNs cannot easily propagate answer-specific information and need to leverage information about the remaining entities to improve the LM's reasoning. The studied setting is illustrated in \Figref{fig:graph_adv}.

We present the results for OBQA and RiddleQA datasets in Table~\ref{tab:findings}. When removing answer information from the KG, GNNs show significant performance degradation and cannot effectively discriminate between correct and incorrect answers. The relative performance degradation is up to 4.8\% for OBQA and using the external KG improves over the LM (w/o KG) by only 0.2\% accuracy points. The experiment suggests that the message passing of GNNs learns to propagate answer-specific information, depends on the connectivity of the answer nodes,  and is limited  when this information is not present or is removed from the graph.

\begin{table}[tb] %
\centering
\caption{QA performance comparison when complete information about candidate answers are in the graph (w/ ans) and when their edges are removed (w/o ans). $\Delta_{\text{acc}}$ denotes the relative performance degradation.}
\label{tab:findings}%
\resizebox{0.7\linewidth}{!}{
\begin{threeparttable}
    \begin{tabular}{l|ccc|ccc}
        \toprule
         & \multicolumn{3}{c|}{OBQA} & \multicolumn{3}{c}{RiddleQA}  \\
        & w/ ans & w/o ans & $\Delta_{\text{acc}}$ &  w/ ans & w/o ans & $\Delta_{\text{acc}}$ \\
        \midrule
        LM (w/o KG) & 64.8 & 64.8 & 0.0 & 60.7 & 60.7 & 0.0\\
        LM + GNN$^*$ & 68.3 & 65.0 & -4.8\% & 66.7 & 64.8 & -2.8\% \\
        \bottomrule
    \end{tabular}%
    
		\begin{tablenotes}
        \item $^*$Results are averaged over three representative QA-GNNs (\Secref{sec:main_results}). The seed LM is \texttt{RoBERTa-Large}.
        \end{tablenotes}
    \end{threeparttable}
}

    \vspace{-0.2in}
\end{table}%

\section{SemPool: Semantic Graph Pooling}
\vspace{-0.1in}
We present a graph-based pooling method, termed SemPool, that aims at robustness during QA with KGs. Unlike message passing methods that depend on \emph{local graph} information around the answer nodes, SemPool leverages \emph{global textual} information from the KG to represent its semantics. Global information is more robust under local graph perturbations, e.g., incomplete edges around nodes, while the textual representation of the KG aids its integration to the LM. SemPool's overall framework is depicted in \Figref{fig:method}. Next, we provide SemPool's components in detail.

\subsection{KG Initialization}
\vspace{-0.1in}
SemPool retrieves a subgraph $\gG_{q_a}$ for context $[c,q,a]$ following existing works~\cite{yasunaga2021qagnn} (\Secref{sec:subgraph}). From now on, we denote the retrieved subgraph as $\gG_q$ (instead of $\gG_{q_a}$) for better readability.
SemPool uses the LM to encode the textual information of each fact $(h,r,t) \in \mathcal{E}_q$ in the retrieved subgraph, aiming at semantic alignment between language and graph, as follows. We transform each edge in the subgraph, $(h,r,t) \in \mathcal{E}_q$, into natural language based on predefined templates for each relation $r \in \gR$. For instance, (\textit{winter, causes, bird migration}) is transformed to ``winter causes bird migration''. Then the verbalized fact, $e$, for edge $(h,r,t)$ is encoded by the \emph{pre-trained} LM. In order to compute a single edge embedding $\vh_e$ for each edge $e$, we use mean-pooling or cls-pooling of the computed token embeddings.

\subsection{Pooling}
\vspace{-0.1in}

SemPool performs global pooling over the edge embeddings $\{ \vh_e : e \in \gE_q \}$ of the retrieved subgraph $\gG_q$.
However, $\gG_q$ is determined based on the linked nodes and their neighbors (\Secref{sec:subgraph}) and as a result, it may contain some noisy facts in the set $\gE_q$. Thus, we propose a self-attention pooling layer that weights the importance of each fact with respect to the  semantics of the subgraph. Specifically, we compute a global graph representation $\vg_q$ with a weighted aggregation by
\begin{equation}
     \vg_q = \sum_{e \in \gE_q} a_{e} f_v( \vh_{e}),    \label{eq:pool}
\end{equation}
where $f_v: \sR^d \xrightarrow{} \sR^d$ is a linear projection and $a_{e} \in [0,1]$ measures the importance of each fact $e \in \gE_q$. Weight $a_{e}$ is computed by a $\softmax(\cdot)$ operation as 
\begin{equation}
    a_{e} = \softmax_{e \in \gE_q} \big( f_k ( \vh_{e}) \big), \label{eq:softmax}
\end{equation}
where $f_k: \sR^d \xrightarrow{} \sR^d$ is a neural network. Learning the importance of each fact facilitates the identification of facts that provide new information to the LM for the QA task.

\begin{figure*}[t]
    \centering
    \includegraphics[width=\linewidth]{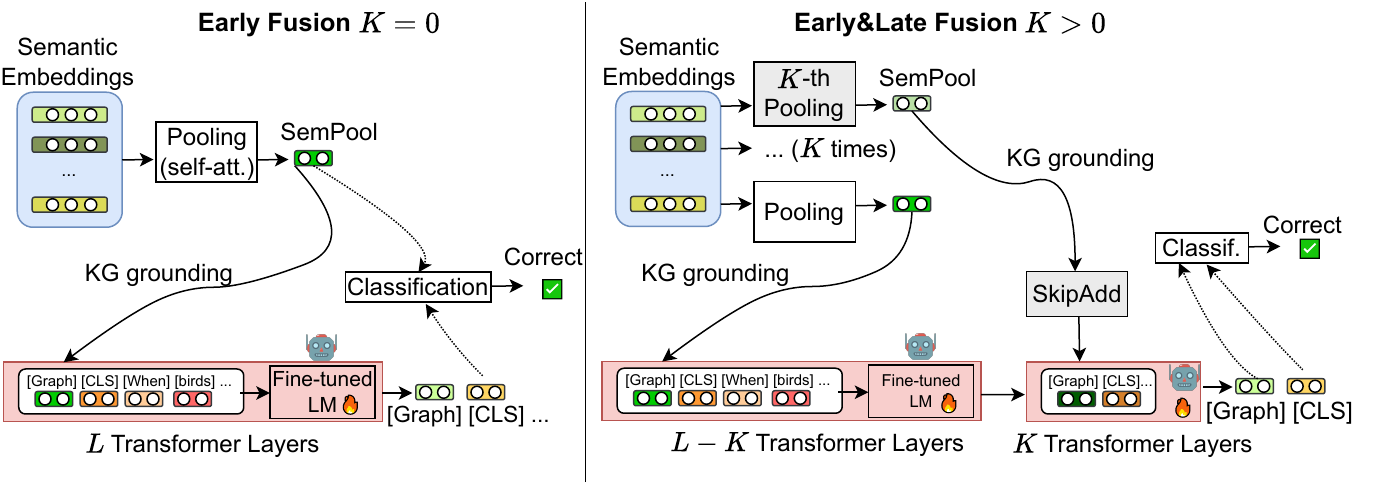}
    \caption{SemPool architecture with early (left) and late (right) fusion. Number $K$ represents the number of late fusion layers. }
    \vspace{-0.2in}
    \label{fig:fusion}
\end{figure*}

\subsection{KG Grounding} \label{sec:fusion}
\vspace{-0.1in}
We ground the LM to the subgraph's semantic information $\vg_q$ by inserting a special \texttt{[Graph]} token in the beginning of the question $q$. The embedding of the \texttt{[Graph]} token is set to $\vg_q$. During training, the LM's transformer layers~\cite{vaswani2017attention} learn to fuse information between the question's tokens and the graph token. 

\textbf{Early Fusion}. In the early fusion approach, the graph representation $\vg_q$ is prepended to the query. After $L$ transformer layers, the LM outputs the final hidden states 
$
    [\vh_{\texttt{Graph}}^{(L)}, \vh_{\texttt{CLS}}^{(L)}, \dots, \vh_T^{(L)}]
$.
We use the $\texttt{[CLS]}$ token as the final question representation $\vq^{(L)}:= \vh_{\texttt{CLS}}^{(L)}$  for answer classification.
Moreover, we use $\vg_q$ for the answer classification loss so that the pooling module gives more attention to facts useful for the QA task. Given an answer candidate $a \in \gA$, its probability $p(a|q)$ of being the correct answer for question $q$ is computed by
\begin{equation}
    p(a|q) = \exp \big( f_q(\vq^{(L)}) + f_g(\vg_q) \big),
\end{equation}
where $f_q, f_g: \sR^d \xrightarrow{} \sR^1$ are MLP networks. During training, we optimize the parameters via the cross entropy loss.

\textbf{Early\&Late Fusion}. The late fusion approach has skip connections~\cite{he2016resent} that fuse graph information into deeper layers of the LM.  This encourages the LM to mix useful graph semantics with language before predictions. The hyperparameter $K$ denotes the $K$ last transformer layers where graph information is fused. For each transformer layer, we have a dedicated pooling module  that computes $\vg_q^{(k)}$, where $k \in \{0, \dots, K\}$. Similar to \Eqref{eq:pool} and \Eqref{eq:softmax}, each $\vg_q^{(k)}$ is obtained via 
\begin{align}
    \vg_q^{(k)} &= \sum_{e \in \gE_q} a_{e}^{(k)} f_v^{(k)}( \vh_{e}), \; \; \;
    a_{e}^{(k)} = \softmax_{e \in \gE_q}( f_k^{(k)} ( \vh_{e}) ).
\end{align}
In the beginning, we set the embedding of the graph token $\vh_g^{(0)}$ to $\vg_q^{(0)}$. At the $(L-k)$-th layer of the LM, $\vh_{\texttt{GRAPH}}^{(L-k)}$ is updated via a skip connection as 
\begin{equation}
    \vh_{\texttt{GRAPH}}^{(L-k)} = \vh_{\texttt{GRAPH}}^{(L-k)} + \vh_g^{(k)}.
\end{equation}
We compute the final answer probabilities as 
\begin{equation}
    p(a|q) = \exp \big( f_q(\vq^{(L)}) + f_g(\vh_{\texttt{GRAPH}}^{(L)}) \big),
\end{equation}
where $f_q, f_g: \sR^d \xrightarrow{} \sR^1$ are MLP networks. Note that the final representation $\vg_q^{(L)}$ used for answer classification depends on the previous states of $\{\vg_q^{(k)}\}_{k=1}^K$, which are optimized altogether during training.

\section{Experimental Setting}
\vspace{-0.1in}
\textbf{QA Datasets}. We evaluate SemPool on three multiple-choice question-answering datasets across two domains. 
OpenBookQA (\textbf{OBQA}; \cite{mihaylov2018obqa}) dataset  is a 4-way multiple-choice QA dataset that requires reasoning with elementary science knowledge. It contains 5,957 questions along with an optional open book of scientific facts. We use the official data split. RiddleSense (\textbf{RiddleQA}; \cite{lin2021riddlesense}) dataset is a 5-way multiple-choice task testing complex riddle-style commonsense reasoning. It has 5,715 questions and we split the dev set in half to make in-house dev/test sets. MedQA-USMLE (\textbf{MedQA}; \cite{jin2021medqa}) dataset is 4-way multiple-choice task that originates from the USMLE practice sets, requiring biomedical and clinical knowledge. The dataset has 12,723 questions and we use the original data splits. 

\begin{table*}[tb] %
\centering
\caption{Test performance comparison on QA datasets. \textcolor{purple}{Purple} color denotes performance degradation at the adversarial setting, while \textcolor{teal}{teal} color denotes improvement.}
\label{tab:main}%
\resizebox{\linewidth}{!}{
\begin{threeparttable}
    \begin{tabular}{l|lllc|lllc|c}
        \toprule
        & \multicolumn{4}{c|}{(w/ ans.)} & \multicolumn{4}{c|}{(w/o ans.)} & \multirow{2}{*}{Avg.}  \\
        & OBQA & RiddleQA & MedQA & Avg. & OBQA & RiddleQA & MedQA & Avg. &\\
        \midrule
        \textit{message passing} & & & & & & & & & \\
        LM + QAGNN & 67.8 ($\pm$2.8)$^*$ & 67.0$^*$ & 38.0$^*$ & 57.60 & 67.0 ($\pm$0.7) & \textcolor{purple}{65.2} ($\pm$0.4) & \textcolor{purple}{36.8} & 56.33 &  56.97 \\
        LM + GreaseLM & 66.9$^*$ & 67.2$^*$ & 38.5$^*$ & 57.53 & \textcolor{purple}{64.4} ($\pm$3.2) & \textcolor{purple}{63.7} ($\pm$1.7) & 39.0 & 55.70 & 56.61 \\
        LM + GSC & 70.3 ($\pm$0.8)$^*$ & 66.0 ($\pm$1.5) & 38.0 & \textbf{58.10} & \textcolor{purple}{63.6} ($\pm$2.6) & 65.6 ($\pm$0.7) & 37.8 & 55.66 & 56.88 \\ 
        \hline
        \textit{no message passing} & & & & & & & & & \\
        LM (w/o KG) &  64.8 ($\pm$2.4)$^*$ & 60.7$^*$ & 37.2$^*$ & 54.23 & 64.8 ($\pm$2.4)$^*$ & 60.7$^*$ & 37.2$^*$ & 54.23 & 54.23\\
        LM + SemPool & 67.7 ($\pm$1.2)  & 67.3 ($\pm$0.4) & 38.9 & 57.96 & \textcolor{teal}{69.5} ($\pm$0.5) & 67.2 ($\pm$1.2)& \textcolor{teal}{39.4} & \textbf{58.70} & \textbf{58.33} \\
        \bottomrule
    \end{tabular}%
		\begin{tablenotes}
        \item $^*$Published results. We use the \texttt{RoBERTa-Large} LM for OBQA and RiddleQA (commonsense). We use the \texttt{SapBERT-Base} LM for MedQA (biomedical).
        \end{tablenotes}
    \end{threeparttable}
}

    \vspace{-0.1in}
\end{table*}%

\textbf{Knowledge Graphs}. Following prior works, we use ConceptNet \cite{speer2017conceptnet}, a general-domain knowledge graph, as our external knowledge source $\gG$ for OBQA and RiddleQA. ConceptNet has 799,273
nodes and 2,487,810 edges in total. For MedQA-USMLE, we use the  KG provided by~\cite{jin2021medqa}. This KG contains 9,958
nodes and 44,561 edges.
For each question, we retrieve  subgraphs following the algorithm of~\cite{yasunaga2021qagnn}. We set the default subgraph size to 32 nodes, which empirically performs well in all datasets. In addition, we study the setting in \Figref{fig:graph_adv}. 

\textbf{Language \& Graph Encoding}. We use (i) RoBERTa-Large~\cite{liu2019roberta} and AristoRoBERTa~\cite{clark2020aristo} for the experiments on OBQA and RiddleQA, and (ii) SapBERT~\cite{liu2021sapbert} and BioLinkBERT-Base~\cite{yasunaga2022linkbert} for  MedQA, demonstrating SemPool's effectiveness with respect to different LM initializations.  We encode KG facts via the respective pretrained LMs for each case: (i) RoBERTa-Large for OBQA and RiddleQA, (ii) SapBERT and BioLinkBERT-Base for  MedQA. %

\textbf{SemPool Implementation}. We follow prior work's implementation for the QA task~\cite{zhang2022greaselm}. We use RAdam optimizer with learning rates selected from $\{1, 2, 5\} \times \mathrm{e}{-5}$ for the LM and set to $1\mathrm{e}{-3}$ for our pooling encoder with a batch size of 32, training epochs are selected from $\{20, 30, 70\}$. For SemPool, we tune $K$ from $\{0,2,3,5\}$ and select cls or mean pooling for the KG edge representation based on the dev set. Experiments are conducted on a Nvidia GeForce RTX 3090 24GB machine.

\textbf{Compared Methods}.
We compare SemPool with representative LM+KG GNN methods: QAGNN~\cite{yasunaga2021qagnn}, GreaseLM~\cite{zhang2022greaselm}, and GSC~\cite{wang2021gsc}. QAGNN uses the question's representation to guide the GNN updates. GreaseLM fuses information from both the language and the graph into the last interaction layers of the LM. QAGNN and GreaseLM use external node embeddings for the KG. GSC treats language and graph separately, and relies on the node/edge types for discriminating between correct and incorrect answers. For a fair comparison, we use the same LM for all compared methods. In addition, we report performance of fine-tuning the LM  without using any KG information, `LM (w/o KG)'.

\section{Results}
\vspace{-0.1in}
\subsection{Main Results} \label{sec:main_results}
\vspace{-0.1in}
\begin{table*}[tb] %
\centering
\caption{Performance comparison on QA datasets with different LMs at the adversarial setting (w/o ans).}
\label{tab:lm_res}%
\resizebox{\linewidth}{!}{
\begin{threeparttable}
    \begin{tabular}{l|cc|cc|cc|c}
        \toprule
        & \multicolumn{2}{c|}{OBQA} & \multicolumn{2}{c|}{RiddleQA} & \multicolumn{2}{c|}{MedQA} & \multirow{2}{*}{Avg.} \\
        & RoBERTa& AristoRoBERTa & RoBERTa& AristoRoBERTa & SapBERT & BioLinkBERT & \\
        & (dev / test) & (dev / test) & (dev / test) & (dev / test) & (dev / test) & (dev / test) & (dev / test) \\
        \midrule
        LM + GNN$^\heartsuit$ & 71.4 / 67.0 & 71.4 / 74.0 & 65.7 / 66.5 & 66.1 / 69.0 &  38.9 / 39.0& 40.7 / 40.9 & 59.03 / 59.40\\
        LM + SemPool & 70.4 / 69.6 & 73.0 / 75.2 & 66.2 / 67.7 &  68.2 / 69.2 & 37.9 / 39.4 & 42.3 / 41.6 & \textbf{59.66} / \textbf{60.45}\\
        \bottomrule
    \end{tabular}%
		\begin{tablenotes}
        \item $^\heartsuit$ We use the best performing GNN from Table~\ref{tab:main}: QAGNN for OBQA, GSC for RiddleQA, GreaseLM for MedQA.
        \end{tablenotes}
    \end{threeparttable}
}

    \vspace{-0.1in}
\end{table*}%

We present the results when comparing SemPool with existing GNN-based (message passing) approaches. Table~\ref{tab:main} shows that SemPool is the most robust method under different configurations and datasets, although it does not involve any complex message passing. SemPool improves over GNNs by 1.45-1.72\% accuracy points on average, while GNNs struggle on the setting when answer node facts are removed from the KG (w/o ans.). %
Moreover, the benefit of grounding the LM's reasoning to the KG becomes clear when comparing SemPool with LM (w/o KG), where SemPool significantly improves performance by 4.1\% points.

In Table~\ref{tab:lm_res}, we present results when using different LMs for the QA task. SemPool outperforms the \emph{best} performing GNN by 0.63\% and 1.05\% accuracy points for the dev and test set, respectively, at the critical setting, when answer information is missing. We observe that SemPool's improvements increase with more powerful LMs: AristoRoBERTa for OBQA and RiddleQA, and BioLinkBERT for MedQA. In these cases, SemPool outperforms the GNNs by 1.6\% (OBQA), 2.1\% (RiddleQA), and 1.7\% (MedQA) accuracy points at the dev set.

\subsection{Ablation Studies \& Analysis} \label{sec:res-abla}
\vspace{-0.1in}
\begin{wraptable}{r}{0.5\textwidth}  %
\centering
 \vspace{-0.4in}
\caption{Dev set performance of different embedding models, averaged over OBQA and RiddleQA datasets.}
\label{tab:obqa_embed}%
\resizebox{\linewidth}{!}{
\begin{threeparttable}
    \vspace{-0.17in}
    \begin{tabular}{l|cc}
        \toprule
        & \multicolumn{2}{c}{\textit{Language Encoder}} \\
        & RoBERTa & AristoRoBERTa \\
        \midrule
        \textit{Graph Encoder} & & \\
        RoBERTa & 68.3 & 70.6 \\
        SBERT & 52.3 & 70.8 \\
        \bottomrule
    \end{tabular}%
    \end{threeparttable}
}
 \vspace{-0.2in}
\end{wraptable}%

\textbf{Semantic Alignment}. Table~\ref{tab:obqa_embed} shows the importance of semantic alignment between language and graph representations. When using RoBERTa for QA and SBERT  for computing graph embeddings, language and graphs semantics are not aligned, which leads to poor performance. On the other hand, using RoBERTa as the graph encoder improves performance by up to 17\% accuracy points. AristoRoBERTa is pre-tuned for QA tasks and thus, it benefits from both RoBERTa and SBERT graph embeddings.

\begin{wrapfigure}{r}{0.5\textwidth}  %
\centering
\vspace{-0.3in}
\resizebox{\linewidth}{!}{
\definecolor{col1}{rgb}{0.60, 0.31, 0.64}
\definecolor{col2}{rgb}{0.30, 0.69, 0.29}
\definecolor{col3}{rgb}{0.22, 0.49, 0.72}
\definecolor{col4}{rgb}{0.89, 0.10, 0.11}
\definecolor{col5}{rgb}{1, 1, 0.8}

\begin{tikzpicture}
\tikzstyle{every node}=[font=\Huge]

\begin{axis}[legend style={at={(.9,0.9),anchor=north east}},
             legend style={legend pos=outer north east,},  title={OBQA},ylabel={Accuracy (\%)}, xlabel={Fusion Layers $K$ }, every axis plot/.append style={ultra thick}, ymin=64, ymax=74, compat=1.5, xtick={0,2,5}
]

\addplot[mark=*,col3, error bars/.cd, y dir=both, y explicit] coordinates {
    (0, 66.6) %
    (2, 70.4) %
    (5, 69.8) %
};

\addplot[mark=square*,col2, error bars/.cd, y dir=both, y explicit] coordinates {
    (0, 72.4) %
    (2, 73.0) %
    (5, 73.0) %
};

\end{axis}

\begin{axis}[legend columns=-1,legend style={at={(0.5,1.4),anchor=north east,}},  title={RiddleQA}, xlabel={Fusion Layers $K$ }, ymin=60, ymax=70,  every axis plot/.append style={ultra thick},  xshift=8cm, compat=1.5, xtick={0,2,5}
]

\addplot[mark=*,col3, error bars/.cd, y dir=both, y explicit] coordinates {
    (0, 65.3) 
    (2, 63.0)
    (5, 66.2)
};
\addlegendentry{roberta-large \; \; } 

\addplot[mark=square*,col2, error bars/.cd, y dir=both, y explicit] coordinates {
    (0, 65.4) %
    (2, 68.2) %
    (5, 67.1) %
};
\addlegendentry{aristo-roberta} 

\end{axis}

\end{tikzpicture}
}
\vspace{-0.2in}
    \caption{Dev set performance with respect to the number $K$ of fusion layers, using two different LMs.}
\label{fig:k_abla}
\vspace{-0.3in}
\end{wrapfigure}
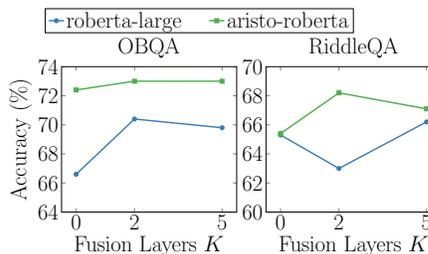

\textbf{Graph Fusion}. \Figref{fig:k_abla} shows the importance of graph to language fusion. In most cases, late fusion ($K>0$) outperforms early fusion ($K=0$) as it injects graph information in multiple LM's layers. The optimal number of fusion layers $K$ is model and task specific, but can be tuned based on the dev set. For example, RiddleQA has more complex questions and requires $K=5$ fusion layers for the RoBERTa LM. 

\begin{figure*}[t]
    \centering
    \includegraphics[width=\linewidth]{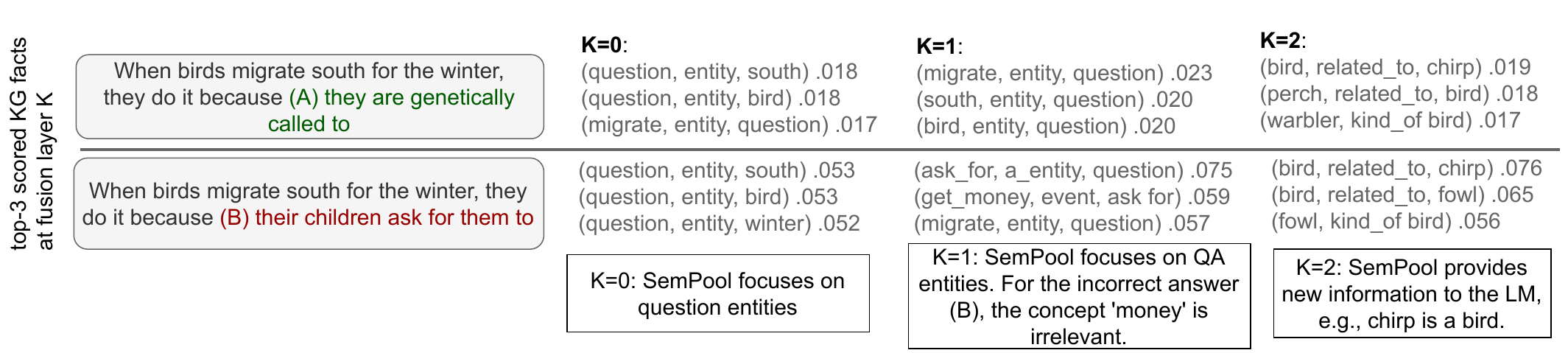}
    \vspace{-0.2in}
    \caption{Working mechanism of SemPool: Top-3 scored facts at each $K \in \{0,1,2\}$, along with their attention weights.}
    \label{fig:analysis}
    \vspace{-0.25in}
\end{figure*}

\textbf{Interpretability}. \Figref{fig:analysis} illustrates the working mechanism of SemPool in one examples case from the OBQA dataset. We observe that different layers of SemPool extract different semantics. At $K=0$, SemPool focuses on question entities (relation: `entity'), which helps the LM  give additional importance to the linked entities during its first layers of reasoning. At $K=1$, SemPool focuses on both question and answer entities (relation: `a\_entity'). Thus, the LM uses additional semantics for the candidate answers. At the last fusion layer, SemPool learns to aggreagate new information for the LM, e.g., \textit{(bird, related\_to, chirp)}. This helps the LM to ground its predictions based on the global KG semantics. For the incorrect answer (B), SemPool identifies the irrelevant concepts `ask\_for' and `get\_money' that do not provide any useful information to the LM.

\section{Conclusions}
\vspace{-0.1in}
We study a critical setting for KG-based QA, where information about the candidate answer entities is missing from the KG. Our empirical results showed that graph-based (message passing) approaches struggle on the QA task under answer-based graph perturbations (\Secref{sec:findings}). We propose SemPool, a graph pooling approach, that is more robust on KG-based QA task as it treats the graph as a set of facts. Experimental results show that SemPool outperforms competing methods by 2.27\% accuracy points, while 
offering interpretability during inference (\Secref{sec:res-abla}). \\

\noindent
\textbf{Acknowledgements}. This work was supported in part by NSF (1447788, 1704074, 1757916, 1834251, 1834332), Army
Research Office (W911NF1810344), Intel Corp, and Amazon Web Services. Access to research and computing facilities was provided by the Minnesota Supercomputing Institute. 
\bibliography{SemPool}
\bibliographystyle{splncs04}

\appendix

Appendix 

\section{SemPool} \label{app:sempool}

\begin{table*}[h] %
\centering
\caption{QA examples from OBQA, RiddleQA, and MedQA datasets.}
\label{app:data-examples}%
\resizebox{\linewidth}{!}{
\begin{threeparttable}
    \begin{tabular}{l|l}
        \toprule
        \multirow{3}{*}{OBQA} & When birds migrate south for the winter, they do it because \\
        & \textcolor{teal}{(A) they are genetically called to}  (B) their children ask for them to \\
        & (C) it is important to their happiness  (D) they decide to each year \\
        \hline
        \multirow{2}{*}{OBQA+fact}  & \underline{Migration is an instinctive behavior.} When birds migrate south for the winter, they do it because \\
        & \textcolor{teal}{(A) they are genetically called to} (B) their children ask for them to \\
        & (C) it is important to their happiness (D) they decide to each year \\
         \hline
         \multirow{1}{*}{RiddleQA} & What turns everything around, but does not move? (A) side \textcolor{teal}{(B) mirror} (C) street corner (D) drive (E) corner \\
         \hline
        \multirow{5}{*}{MedQA} & A 51-year-old female presents with intermittent right upper quadrant discomfort. The physician suspects \\ 
        & she is suffering from biliary colic and recommends surgery. Following surgery, brown stones are removed from  \\
        & the gallbladder specimen.  What is the most likely cause of the gallstone coloring? \\
        & \textcolor{teal}{(A) E. coli infection; beta-glucoronidase release} (B) Shigella infection; HMG-CoA reductase release \\
        & (C) Shigella infection; beta-glucoronidase release (D) Bile supersaturated with cholesterol; beta-glucoronidase release \\
        \bottomrule
    \end{tabular}%
        
    \end{threeparttable}
}

    \vspace{-0.1in}
\end{table*}%

\subsection{Comparison with Existing Approaches}

\textbf{KG Grounding}.
One benefit of SemPool is that KG information is inserted into the LM, grounding its reasoning at different layers. Most existing approaches do not fuse information into the LM's layers~\cite{yasunaga2021qagnn,wang2021gsc} or fuse information at last layers only~\cite{zhang2022greaselm,sun2021jointlk,park2023qat}. The goal of these methods is to use the question representation to guide the GNN updates. On the other hand, SemPool aims at improving the LM's reasoning, by conditioning its transformer layers to KG information. 

\noindent
\textbf{Semantic Alignment}. 
SemPool uses the same seed LM to encode textual information from the KG facts as well as for initializing the QA-finetuned LM. As a result, SemPool represents both the graph and the language in the same semantic space, which can facilitate knowledge exchange between the two modalities. In contrast, most existing approaches~\cite{feng2020scalable} use external embeddings to represent the graph, which might require additional learning for aligning the graph and the language representations.

\subsection{Complexity}

 GNNs are \emph{node-centric} and they recursively update the node embeddings at different layers (\Secref{sec:gnn}). For each node in $\gV_q$, GNNs perform $\gO(K)$ aggregations, where $K$ is the number of GNN layers, requiring $\gO(|\gV_q| K)$ total aggregations. At each aggregation, GNNs aggregate $\gO(\Delta)$ messages, where $\Delta$ is the maximum node degree (usually, $\Delta \ll  |\gV_q|$ in sparse graphs).
SemPool is \emph{graph-centric} as it pre-computes \emph{edge} embeddings and aggregates them once into a single representation during inference. SemPool only requires $\gO(K+1)$ \emph{total} aggregations, where $K$ is the number of fusion layers. Each aggregation involves $|\gE_q|$ pre-computed messages.

\section{Datasets \& Experimental Setting} \label{app:data-exp}
We provide example cases of the QA datasets used in Table~\ref{app:data-examples}. Hyperparameter settings for the experiments are given in Table~\ref{tab:app-parameters}.

\begin{table}[tb] %
\centering
\caption{Hyperparameter settings for experiments.}
\label{tab:app-parameters}%
\resizebox{0.7\linewidth}{!}{
\begin{threeparttable}
    \begin{tabular}{l|cccc}
        \toprule
        Category & Hyperparameter & OBQA / RiddleQA & MedQA \\
        \midrule
         \multirow{2}{*}{Model} & Number of fusion layers & $\{0, 2, 5\}$ & $\{0, 3\}$  \\
         & Token pooling & \{cls, mean\} & \{cls, mean\} \\
         \hline 
        \multirow{5}{*}{Optimization} & LM learning rate & $1e^{-5}$ & $\{2e^{-5}, 5e^{-5}\}$ \\
         & Graph encoder learning rate & $1e^{-3}$ & $1e^{-3}$ \\
         & Optimizer & RAdam & RAdam \\
         & Epochs & 70 / 30 & 20 \\
         & Batch size & 32 & 32 \\
         \hline
         \multirow{2}{*}{Data} & Max number of nodes & 32 & 32 \\
         & Max number of tokens & 100 & 512 \\
        \bottomrule
    \end{tabular}%
    \end{threeparttable}
}

\end{table}%

\section{Results \& Ablation Studies}

\subsection{Subgraph Size}
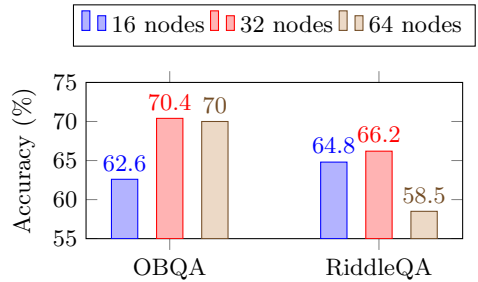
\begin{figure*}[tb]
\centering
  \begin{subfigure}[b]{0.6\linewidth}
  \pgfplotstableread[row sep=\\,col sep=&]{
Dataset & A  & B & C \\
OBQA    & 62.6  & 70.4  & 70.0 \\
RiddleQA   & 64.8  & 66.2  & 58.5 \\
}\graphdata

\begin{tikzpicture}
\tikzstyle{every node}=[font=\footnotesize]
    \begin{axis}[
            ybar=.25cm,
            bar width=.35cm,
            enlarge x limits={0.4},
            width=.9\columnwidth,
            height=0.5\columnwidth,
            legend style={at={(.5,1.5)},
                anchor=north,legend columns=-1},
            symbolic x coords={OBQA, RiddleQA},
            xtick=data,
            nodes near coords,
            nodes near coords align={vertical},
            ymin=55,ymax=75,
            ylabel={Accuracy (\%)},
            ylabel near ticks,
            compat=1.5, 
        ]
	\addplot table[x=Dataset,y=A]{\graphdata};
        \addplot table[x=Dataset,y=B]{\graphdata};
        \addplot table[x=Dataset,y=C]{\graphdata};
        \legend{16 nodes, 32 nodes, 64 nodes}
    \end{axis}

\end{tikzpicture}
  \caption{Dev set performance with respect to the subgraph size, setting the maximum node number to $\{16, 32, 64\}$.}
    \label{fig:graph}
  \end{subfigure}
  \caption{Ablation studies of different SemPool's components. %
  }
\end{figure*}

In \Figref{fig:graph}, we study the effect of the subgraph size retrieved, setting the number of maximum nodes to $\{16, 32, 64\}$. For OBQA, the retrieved subgraphs are sparser with 110-118 nodes on average, while for RiddleQA,  the retrieved subgraphs are denser with 167-200 nodes on average. As \Figref{fig:graph} shows, OBQA benefits from subgraphs with more edges that can provide additional factual information ($|\gV_q| \in \{32, 64\}$), while RiddleQA benefits from smaller graphs that include fewer noisy edges ($|\gV_q| \in \{16, 32\}$).

\subsection{Integrating SemPool to QA systems} \label{sec:sempool-sota}

\begin{wraptable}{r}{0.4\textwidth} %
\centering
\vspace{-0.1in}
\caption{SOTA performance on OBQA using additional scientific text.}
\label{tab:obqa}%
\resizebox{0.7\linewidth}{!}{
\begin{threeparttable}
    \vspace{-0.2in}
    \begin{tabular}{l|c}
        \toprule
        System (\#Params) & Acc.  \\
        \midrule
        T5 (3B) \cite{raffel2020t5} & 83.2  \\
        T5+KB ($\geq$11B) & 85.4 \\
        UnifiedQA (11B) \cite{khashabi2020unifiedqa} & 87.2  \\
        \hline 
        GreaseLM (359M) \cite{zhang2022greaselm} & 84.8 \\
        DRAGON (359M) \cite{yasunaga2022dragon} & 87.8  \\
        \midrule
        AristoRoBERTa (355M) \cite{clark2020aristo} & 77.8 \\
        \; + QAGNN \cite{yasunaga2021qagnn} & 82.8\\
        \; + JointLK \cite{sun2021jointlk} & 85.6\\
        \; + GSC \cite{wang2021gsc} & 87.4\\
        \; + QAT  \cite{park2023qat} & 87.6\\
        \; + SemPool + GSC (\textbf{ours}) & \textbf{88.2}\\
        \bottomrule
    \end{tabular}%
    \end{threeparttable}
}
\vspace{-0.3in}
\end{wraptable}%

We further experiment using AristoRoBERTa with additional text data for OBQA, similar to SOTA systems. Table shows the potential of SemPool: Grounding the LM's reasoning to the KG (SemPool) and combining it with message passing over the KG (GSC), we can achieve higher accuracy than billion-scale LMs or SOTA systems for KG powered QA, such as DRAGON and QAT.

\end{document}